\documentclass[letterpaper, 10 pt, conference]{ieeeconf}  

\IEEEoverridecommandlockouts                              

\usepackage[dvipsnames]{xcolor}

\definecolor{gqorange}{HTML}{EE4431}

\definecolor{hcblue}{HTML}{1E90FF}

\definecolor{cvprblue}{rgb}{0.12,0.49,0.85}

\usepackage{booktabs}
\usepackage{pifont}
\usepackage{xcolor}
\usepackage{caption}
\usepackage{graphicx} 
\usepackage{amsmath}  
\usepackage{etoolbox} 
\usepackage{multirow}
\usepackage{wrapfig}
\usepackage{listings}

\usepackage{xspace}

\newcommand{\acronym}{GSWorld}

\usepackage{xcolor}
\usepackage{booktabs}
\usepackage{epsfig}
\usepackage{graphicx}
\usepackage{amsmath}
\usepackage{amssymb}
\usepackage{subcaption}
\usepackage{pifont}
\usepackage{wrapfig}
\let\labelindent\relax
\usepackage{enumitem}
\usepackage[breaklinks,colorlinks = true,citecolor=cvprblue, urlcolor=cvprblue]{hyperref}
\usepackage{cleveref}

\usepackage{dsfont}
\usepackage{bbm}
\usepackage{algorithm}
\usepackage{algpseudocode}

\usepackage{pifont}
\usepackage{bm}

\usepackage
[sortcites,
backend=bibtex,
bibstyle=ieee,
citestyle=numeric,
sorting = nyt,
mincitenames=1,
maxcitenames=2,
natbib=true,
doi=false,
isbn=false,
url=false,
eprint=false]{biblatex}
\title{\acronym{}: Closed-Loop Photo-Realistic Simulation Suite \\ for Robotic Manipulation}

\author{
Guangqi Jiang$^{*1}$ \quad Haoran Chang$^{*1,2}$ \quad Ri-Zhao Qiu$^{1}$ \quad Yutong Liang$^{1}$ \quad Mazeyu Ji$^{1}$ \quad Jiyue Zhu$^{1}$ \\ Zhao Dong$^{3}$ \quad Xueyan Zou$^{1}$ \quad Xiaolong Wang$^{1}$
\\ $^{1}$UC San Diego, $^{2}$UC Los Angeles, $^{3}$Meta  \\ *Equal Contributions \quad \href{https://3dgsworld.github.io}{https://3dgsworld.github.io} 
}

\begin{document}

\makeatletter
\let\@oldmaketitle\@maketitle
    \renewcommand{\@maketitle}{\@oldmaketitle
    \centering
    \includegraphics[width=1.0\textwidth]{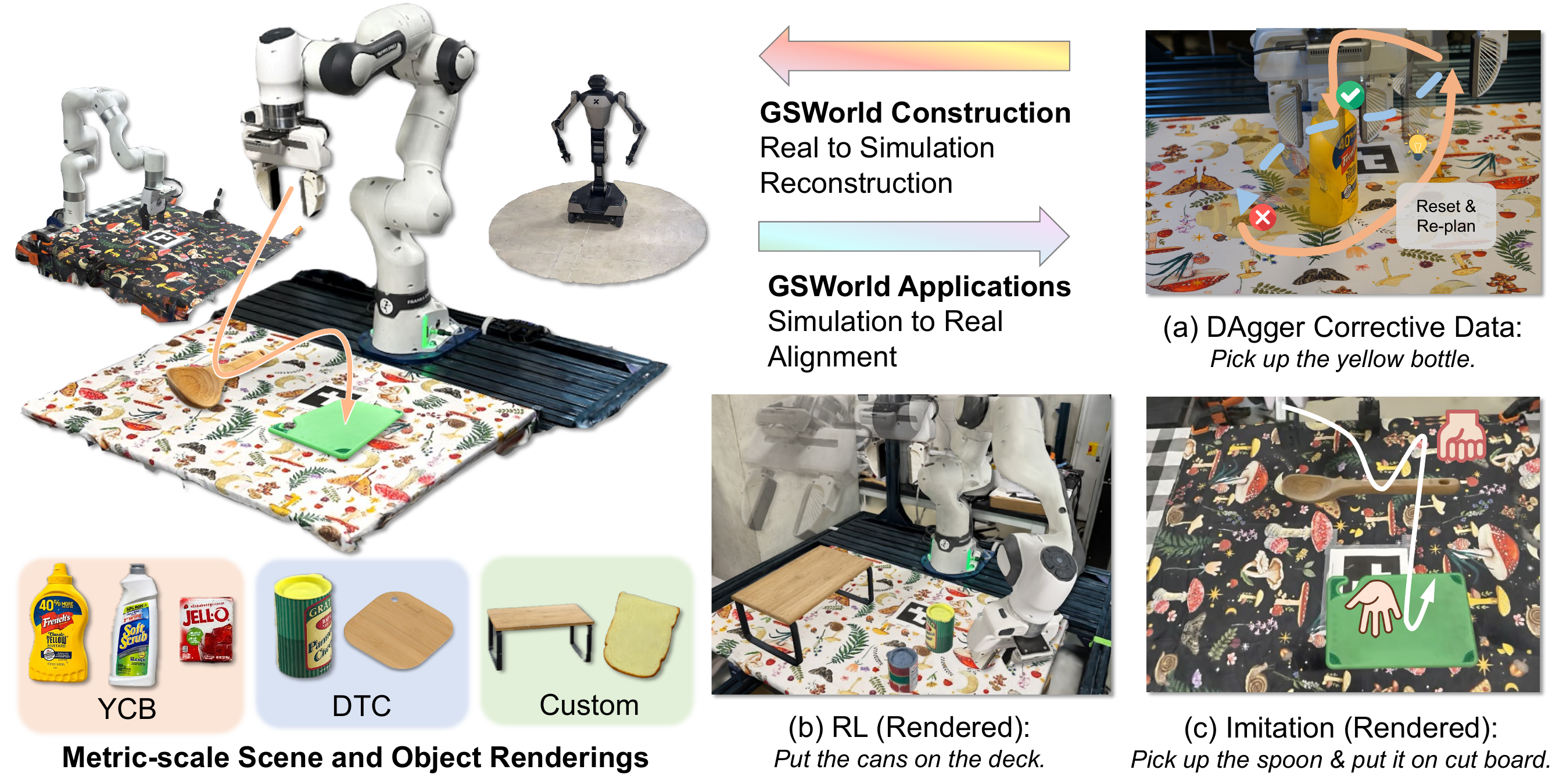}
    \captionof{figure}{\acronym{} leverages 3DGS reconstruction to render photo-realistic robot scenes, supports multiple policy learning recipes in the simulation, and realizes zero-shot sim2real transfer. \acronym{} also applies to policy visual benchmarking and virtual teleoperation for data collection.}
    \vspace{-0.2cm}
    \label{fig:teaser}
    \setcounter{figure}{1}
  }
\makeatother

\maketitle

\thispagestyle{empty}
\pagestyle{empty}

\begin{abstract}
    This paper presents \acronym{}, a robust, photo-realistic simulator for robotics manipulation that combines 3D Gaussian Splatting with physics engines. Our framework advocates `closing the loop' of developing manipulation policies with \textbf{reproducible evaluation} of policies learned from real-robot data and \textbf{sim2real policy training} without using real robots.
    To enable photo-realistic rendering of diverse scenes, we propose a new asset format, which we term GSDF (Gaussian Scene Description File), that infuses Gaussian-on-Mesh representation with robot URDF and other objects. With a streamlined reconstruction pipeline, we curate a database of GSDF that contains 3 robot embodiments for single-arm and bimanual manipulation, as well as more than 40 objects.
    Combining GSDF with physics engines, we demonstrate several immediate interesting applications: (1) learning zero-shot sim2real pixel-to-action manipulation policy with photo-realistic rendering, (2) automated high-quality DAgger data collection for adapting policies to deployment environments, (3) reproducible benchmarking of real-robot manipulation policies in simulation, (4) simulation data collection by virtual teleoperation, and (5) zero-shot sim2real visual reinforcement learning. 
\end{abstract}
\section{Introduction}

Training manipulation policies typically relies on three data sources—simulation, human videos, and real teleoperation—each presenting a distinct trade-off. While simulation provides a perfectly aligned action space for the robot, it often suffers from many sim-to-real gaps. Human videos offer the benefit of photo-realistic scenes and real physics, but lack temporally aligned robot actions and operate in a mismatched action space. Teleoperation successfully aligns both perception and actions, yet its high cost and difficulty to scale are significant limitations. 

To resolve these trade-offs, we introduce \acronym{}, a closed-loop, photo-realistic simulation suite that couples 3D Gaussian Splatting (3DGS) with physics to narrow both the visual and action-space gaps for manipulation. “Closed-loop” here means the same environment can be used to train, evaluate, diagnose failures, and relabel, enabling rapid iteration: policies perceive photo-realistic renderings while issuing controls in the robot’s native space, all inside a simulator that mirrors real scenes closely enough to support zero-shot transfer and efficient adaptation. We demonstrate that \acronym{} enables a range of downstream applications and, in particular, effective sim-to-real transfer for imitation learning, reinforcement learning, and DAgger-style data collection.

This closed-loop capability is powered by a bidirectional pipeline that ensures tight alignment between the physical world and its digital twin. In the \textbf{\textit{real-to-sim}} direction, our pipeline reconstructs a metric-accurate digital twin from short multi-view captures, sets an absolute scale with ArUco markers, and aligns the robot URDF to the scene via surface fitting (e.g., ICP). We then attach collision meshes and material properties to produce a versatile GSDF asset. The \textbf{\textit{sim-to-real}} direction is the reverse: policies trained in \acronym{} deploy on hardware without interface translation because their control and observation spaces match the robot’s native APIs. Policies trained with both sim and real data can be deployed in the sim to evaluate, detect failures, and gather DAgger corrections, thus closing the iteration loop. Achieving this sim-to-real alignment requires that the scene’s geometry, camera properties, and action semantics remain consistent across the real-to-sim divide. We measure the degree of this correspondence through a suite of metrics evaluating the visual, geometric, and functional similarity between the two worlds.

With photo-realistic perception and native action-space control in one loop, \acronym{} supports zero-shot sim-to-real for both visual imitation learning and visual RL, while exploiting scalable parallelism to accelerate data generation and training. Its closed-loop DAgger workflow lets practitioners reproduce on-robot failures inside the digital twin, step through them frame-by-frame, and collect targeted corrective labels with far less teleoperation overhead. Finally, \acronym{} provides reproducible visual benchmarking: shared GSDF assets, fixed camera intrinsics/extrinsics, consistent lighting/materials, and standardized action semantics enable apples-to-apples comparisons across robots, scenes, and tasks—so improvements reflect algorithmic progress rather than environmental variance.

Existing GS-based simulators either target single-setup photo-realistic rendering, provide engine-tied pipelines without a portable asset standard, or limit the reproducible cross-embodiment benchmarking and deployment-oriented on-policy data collection \cite{qureshi2025splatsim,li2024robogsim,Mu_2025_CVPR,chen2025robotwin}. While \acronym{} delivers an effective real-to-sim-to-real workflow that unifies photo-realistic 3DGS with contact-accurate physics, enabling scalable cross-embodiment benchmarking, zero-shot imitation and reinforcement learning, and automated high-quality DAgger data collection for continual deployment-time improvement.

In summary, our contributions are:
\begin{itemize}[leftmargin=3mm]
    \item {A solid real-to-sim-to-real pipeline. Our robust real-to-sim-to-real pipeline accurately aligns the simulation with the real environment, enabling a wide range of subsequent applications.}
    \item {Simulation Data Collection and Visual Imitation Learning (IL).} \acronym{} supports multiple sim data collection methods, \textit{e.g.} motion planning, teleoperation. IL policies trained with \acronym{} data can be directly deployed to the reconstructed real-world scenes.
    \item {Visual RL.} \acronym{} is designed to utilize parallel environments in the simulation to train RL policies. We provide an analysis to show that \acronym{} reduces RL sim2real visual gaps.
    \item {Closed-loop DAgger Learning with Visual Benchmarking.} \acronym{} shows reliable policy evaluation results in correlation with real-world deployments, contributing to using DAgger to iteratively improve real-world policies.
\end{itemize}

\begin{figure*}[h!]
  \centering
  \includegraphics[width=1.\linewidth]{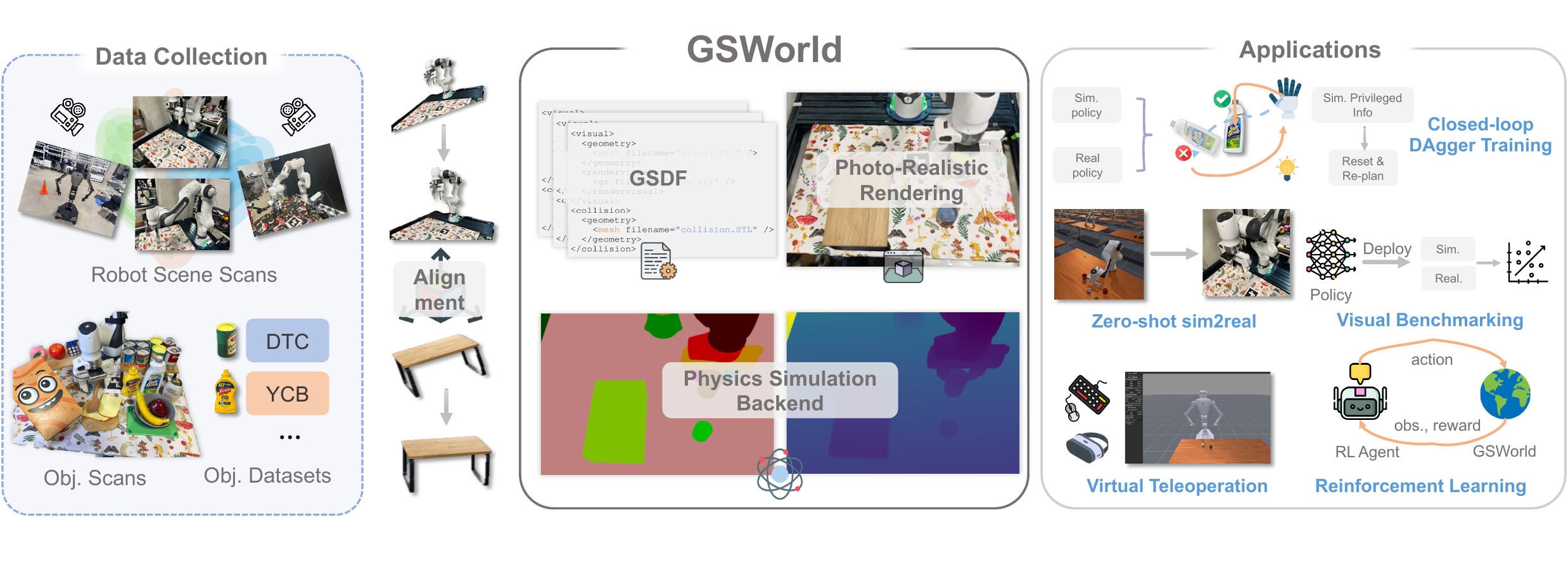}
  \vspace{-30pt}
  \caption{\acronym{} provides an interface on top of existing simulators to render photorealistic assets. Our GSDF assets are compatible with existing simulators to use standard formats for rendering visuals ({\it e.g., depth, segmentation}) and computing physics collisions. \acronym{} provides a rendering wrapper on top of simulators to make the RGB rendering photo-realistic to support various domain randomization and applications.}
  \label{fig:method_overview}
  \vspace{-12pt}
\end{figure*}
\section{Related Work}

\textbf{Robotics Simulation.}
With developments in computer graphics for rendering~\cite{whitted1979-raytracing} and object material simulation~\cite{sulsky1994-MPM}, the robotics community has designed various physics engines~\cite{coumans2016-pybullet,erez2015-physX} and simulators~\cite{xiang2020sapien,rohmer2013-CoppeliaSim,zakka2025-mujocojax} to support various robotics tasks. Recently, more and more simulators have started to improve efficiency and fidelity. For example, Mujoco-Jax~\cite{zakka2025-mujocojax} exploits just-in-time compilers~\cite{hu2019-taichi} to achieve impressive simulation efficiency in Python. To improve rendering fidelity and reduce \textbf{sim2real visual gaps}, recent papers have turned to sophisticated ray-tracing techniques~\cite {whitted1979-raytracing} and generative AI tools~\cite{nasiriany2024-robocasa} to reduce the visual observation gap between simulation renderings and the real world. A recent `meta-simulator', Roboverse~\cite{geng2025-roboverse}, attempts to provides a unified simulation interface to make use of the advantages in individual simulators. This paper focuses on improving the fidelity of simulation renderings by combining recent advances in 3D Gaussian Splatting and simulators.

\textbf{Real2Sim (Real-to-Simulation).}
As an alternative to modeling basic physics and elements from the bottom-up in simulation, real2sim approaches take a different approach to build simulation assets by virtualizing real-world assets. Recent real2sim methods can be roughly divided into two categories: photo-realistic 3D reconstruction~\cite{weng2024-digital-twin-art,qu2024livescene,chen2024omnire,qureshi2025splatsim,li2024robogsim,lou2024-robogs,han2025-re3sim,chakra2024-embodiedGS,lu2024-manigaussian} and part-level (articulated) object understanding~\cite{dai2024-ACDC,mandi2024real2code,chen2024-urdformer,torne2024-rialto}.

\textbf{Real2Sim (Real-to-Simulation) - Photo-realistic Rendering.}
Early method in Real2Sim reconstruction~\cite{qu2024livescene} use NeRFs~\cite{mildenhall2021nerf}, Diffusion~\cite{li24simpler} or Mesh~\cite{torne2024-rialto} for photo-realistic modeling. Due to the inherent implicit representation of NeRFs, they often rely on surrogates such as deformation fields~\cite{jambon2023-nerfshop} to deform the visual renderings to accommodate object motion, which is unnatural and inefficient. On the other hand, mesh-based representations contain many artifacts~\cite{torne2024-rialto} that lead to visual gap. Simpler~\cite{li24simpler} proposes to use generative modeling to advance reproducible benchmarking~\cite{james2020-rlbench,jiang2024robots} by supporting photorealistic rendering. However, Simpler requires expensive manual labor to match green screen and textures, hindering its scalability. With developments in 3DGS~\cite{kerbl20233d}, a rasterization-based method, SplatSim~\cite{qureshi2025splatsim} successfully combined 3DGS with PyBullet~\cite{coumans2016-pybullet} to build a photo-realistic simulator and demonstrated zero-shot sim2real policy deployment. As 3DGS can be explicitly represented as `Gaussian Blobs', the photo-realistic appearance can be displaced consistently with object physics. However, SplatSim relies on manual 3D segmentation of both the robot arm and objects, which is overfitted to a single scene. In parallel to the development of SplatSim, Embodied-GS~\cite{chakra2024-embodiedGS} learns arm-object interaction without using physics engine in a single scene; Robo-GS~\cite{lou2024-robogs} focuses on identifying physical parameters of individual objects from rendering; and ManiGaussian~\cite{lu2024-manigaussian} investigates optimizing Gaussian representations from simulators with ideally synchronized multi-view information. Most recently, Re3Sim~\cite{han2025-re3sim} extends SplatSim~\cite{qureshi2025splatsim} and found that the capability to perform photo-realistic simulation leads to more robust manipulation policies with domain randomization and mixed simulation ({\it e.g.,} non-photorealistic simulation assets with photo-realistic scenes). We continue to advance the progress in photo-realistic simulation by combining the latest advancements in 3DGS and introducing more 3DGS assets with a unified asset format.

\textbf{Real2Sim (Real-to-Simulation) - Others.} Here we provide a concise review of other progress in real2sim that is orthogonal to our method. Part-level (articulated) object understanding methods~\cite{dai2024-ACDC,mandi2024real2code,chen2024-urdformer,torne2024-rialto} apply internet-scale pre-trained visual and language models~\cite{radford2021-clip,liu2024-groundingdino} to create physics and articulation of simulation assets from real-world observations. While these methods focus on understanding articulation of objects, recent methods have also worked on more challenging tasks such as estimating physics for deformable objects~\cite{jiang2025-phystwin} and rigid contracts~\cite{pfaff2025-scalable-real2sim}. PhysTwin~\cite{jiang2025-phystwin} optimizes physics models for elastic objects by assuming a Spring-Mass model and estimating physics parameters from video observations. Scalable Real2Sim~\cite{pfaff2025-scalable-real2sim} is an advancement from previous physics estimation method~\cite{lou2024-robogs}, in which the authors built a pipeline that uses robot arm and camera setups to automate estimation of rigid physics parameters including mass, center of mass, and inertia tensor~\cite{pfaff2025-scalable-real2sim}. \citet{liu2024-diff-rendering} proposes to optimize robot kinematics from differentiable rendering.

\vspace{-2pt}
\section{Method}

\vspace{-5pt}
\subsection{Problem Formulation}

We consider the problem of learning robot policies from visual observations. Let $\mathcal{S}_{\text{sim}}$ denote simulated scene and $\mathcal{G}_{\text{real}}$ denote the real scene $\mathcal{S}_{\text{real}}$ reconstructed by 3DGS from multiple RGB viewpoints $\mathcal{V} = \Sigma v_{i}$. $\mathcal{G}_{real}$ can be used to render novel-view RGB images $I^{\text{gs}} = \mathcal{G}_{real}(p, s)$, which enables photorealistic rendering of the scene under arbitrary camera poses $p$ and environment states $s$.

Our goal is to replace raw real-world RGB observations $I^{\text{real}}$ with 3DGS-rendered images $I^{\text{gs}}$ for downstream robot learning tasks, including IL, RL, and DAgger. 
Formally, at each time step $t$, the underlying state of the system is represented as
\vspace{-2pt}
\begin{equation}
s_t = (q_t, x_t^1, \dots, x_t^n),
\end{equation}
where $q_t \in \mathbb{R}^m$ denotes the robot’s joint position, and $x_t^k$ represents the 6D pose of the $k$-th object in $\mathcal{S}_{\text{real}}$. We mainly use joint position control for the robots.
The robot receives an observation 
\vspace{-4pt}
\begin{equation}
o_t = I^{\text{gs}}_t = \mathcal{G}_{real}(p_t, s_t),
\end{equation}
\vspace{-2pt}
rendered from $\mathcal{G}_{\text{real}}$ given the current camera pose $p_t$ and environment state $s_t$.

In RL, the policy $\pi_\theta$ is trained by interacting with the environment and receiving rewards. 
In IL, the expert $\mathcal{E}$ provides demonstrations $\tau_\mathcal{E} = \{(q_1, o_1,a_1), \dots, (q_T, o_T,a_T)\}$, which are used to supervise $\pi_\theta$. 
In DAgger, $\pi_\theta$ is iteratively refined by collecting expert rollouts from previous failure cases, getting $\tau_\mathcal{D}$. 
In all cases, the policy takes as input $I^{\text{gs}}_t$ instead of $I^{\text{real}}_t$, and robot proprioceptions $q_t$, i.e.,
\vspace{-2pt}
\begin{equation}
a_t \sim \pi_\theta(I^{\text{gs}}_t, q_t).
\end{equation}

\vspace{-2pt}At test time, the trained policy $\pi_\theta$ must generalize to real-world observations $I^{\text{real}}$, ensuring that \acronym{} bridges sim2real visual gap.

\subsection{Recipe for Real2Sim Reconstruction}

This section describes how \acronym{} creates GSDF assets to construct photo-realistic scenes. Compared with existing work~\cite{qureshi2025splatsim,lou2024-robogs,li2024robogsim,han2025-re3sim, yang2025-novel} that focus on constructing robots in a single scene, \acronym{} is designed (1) with an easy-to-use streamlined procedure to reduce manual alignment efforts and (2) to incorporate recent advancements in 3DGS such as geometric accuracy~\cite{huang2024-2dgs}.

\paragraph{Collecting Training Views} To construct a scene with a robot, we use both robot sensors ({\it i.e.,} wrist cameras and third-person cameras) and mobile phone cameras while saving the current joint pose of robot during the scene capture.

\paragraph{Aligning Scale for Metric Representation} Existing real2sim2real methods~\cite{qiu2024-featuresplatting, qureshi2025splatsim} relies on COLMAP~\cite{schonberger2016-colmap}, which introduces scale ambiguity. While such ambiguity can be dealt with manually for a single scene, it affects the scalability for multiple robot embodiments and scenes. To avoid manual scale alignment, we use a simple solution to include a printed ArUco marker~\cite{garrido2014-aruco, ji2024-graspsplats} on the tabletop during the data collection (qualitative examples in Fig.~\ref{fig:teaser}). The detected keypoints of ArUcO markers are projected onto the point cloud formed by 3DGS. We then scale the point cloud using the known scales of the ArUcO marker. In addition, the ArUco marker helps identify the support surface in collision and estimate gravity direction.

\paragraph{Aligning Robots and Table} Given a metric-scale $\mathcal{G}_{real}$ of a static robot $\mathcal{R}$ and a metric-scale robot URDF in $\mathcal{S}_{sim}$, we align the simulation joint position with the real world. Then, we sample and densify surface point clouds from the visual mesh of the robot URDF. We then perform an ICP to compute the rigid transform $\mathcal{T}_{\mathcal{R}, sim}^{gs}$: $\mathcal{G}_{real} = \mathcal{T}_{\mathcal{R}, sim}^{gs} \cdot \mathcal{S}_{sim}$. Compared to previous methods~\cite{qureshi2025splatsim}, our ICP has fewer degrees of freedom since the scale is fixed. Given the alignment, we use K-NN to segment robot links in $\mathcal{G}_{real}$.

\paragraph{Object Assets} Our prior reconstruction stage focuses on background and robot scans. For moveable objects $\mathcal{O}$, we consider integrating existing large-scale datasets and supporting custom objects for generalizability. Specifically, we use DTC~\cite{dong2025-dtc} for its photo-realistic visual quality, and YCB~\cite{calli2017yale}. For custom objects, we use 2DGS~\cite{huang2024-2dgs} to get the reconstruction $\mathcal{O}^{gs}$ and mesh reconstruction. Mass is estimated by weighing. The unobserved bottom regions of the object can be optionally inpainted using amodal reconstruction~\cite{wu2025-amodal3r,agnew2021amodal} or 3D object generation~\cite{liu2023zero} method. Similarly, we use ICP to get the transform $\mathcal{T}_{k, sim}^{gs}$ for the $k$-th object: $\mathcal{O}_k^{gs} = \mathcal{T}_{k, sim}^{gs} \cdot \mathcal{O}_k$.

\subsection{Applications - Closing the Loop for Developing Visual Manipulation Policies}

\paragraph{Closed-loop DAgger Training}
In the traditional imitation learning settings, policy weights are not updated once they are trained and deployed. During deployment, policies often run into failures. DAgger~\cite{ross2011-dagger} is a solution to this case where corrective data is used to train the model to adapt to failure cases. DAgger data has been shown to have much better data efficiency than plain data by previous works~\cite{liu2022-humaninloopdagger,zhang2024-diffusiondagger}. However, collecting DAgger data is hard as it requires reproducing scene setups for the model to `re-experience' the failure case.

\begin{figure}[ht]
    \vspace{-10pt}
    \centering
    \includegraphics[width=0.98\linewidth]{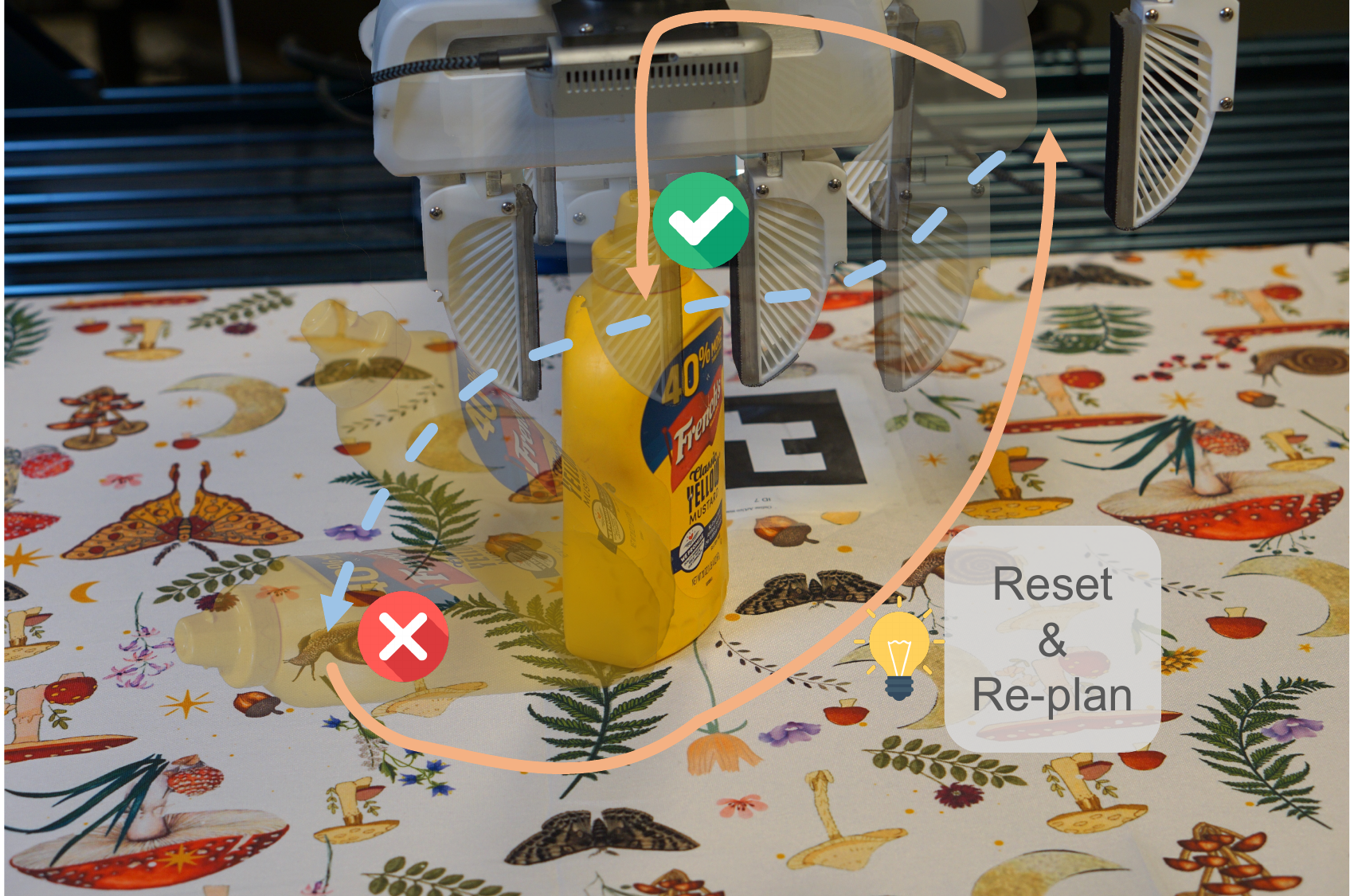}
    \caption{\small{\textbf{DAgger Data Collection in Simulation}. With privileged information provided by simulation, we can record and relive the failure cases during rollouts and generate corrective data for policy adaptation.}}
    \label{fig:dagger-idea}
    \vspace{-6pt}
\end{figure}

\acronym{} provides an interface for automatic DAgger data collection in simulation. Given the GSDF of the target deployment environment and tasks with scripted policies, we roll out the policies with \acronym{}. For failure recordings $\mathcal{D}_f = (s_1, \dots, s_T)$, we randomly sample recovery states $s_r \sim \mathcal{D}_f$ with a uniform sampler, where the task is still achievable in $s_r$, and run the motion planner to obtain corrective data, as illustrated in Fig.~\ref{fig:dagger-idea}. 

To iteratively improve zero-shot sim2real policies, we collect data, evaluate policies, record failures, and recover, resulting in an aggregated dataset $\tau_\mathcal{S}$:
\vspace{-2pt}
\begin{equation}
\tau_\mathcal{S} = \Sigma_i{(\mathcal{Q}_s, \mathcal{O}_s, \mathcal{A}_s)_i}
\end{equation}
\vspace{-2pt}
, where $\mathcal{Q}_s, \mathcal{A}_s$ denote simulation robot joint positions, and action labels. $\mathcal{O}_s = I^{gs}$ is the \acronym{}-rendered observations and $i$ is the DAgger iteration.

To improve policies trained with real-world data, we first evaluate the polices and repeat the same loop, leading to dataset $\tau_\mathcal{R}$:
\vspace{-2pt}
\begin{equation}
\tau_\mathcal{R} = (\mathcal{Q}_r, \mathcal{O}_r, \mathcal{A}_r) \cup \tau_\mathcal{S}
\end{equation}
\vspace{-2pt}
, where $\mathcal{Q}_r, \mathcal{O}_r, \mathcal{A}_r$ are collected in the real world by teleoperation.

\paragraph{Visual Benchmarking.} Recent VLAs~\cite{black2024-pi0,liu2024-rdt,ghosh2024-octo} train generalist policies that work on many different robot embodiments. Learning from extensive cross-embodiment training data, these base models adapt to unseen robot hardware and language instructions in a few-shot or even zero-shot manner~\cite{black2024-pi0,liu2024-rdt}. However, due to their reliance on real robot data during training, there does not exist a standardized visual manipulation benchmark that studies the quality of the base models and their data sampling efficiency for novel embodiments.

\acronym{} provides a photo-realistic rendering interface to address such an issue. With our GSDF assets, we provide photo-realistic simulation of various scenes with different robot embodiments, and a wide selection of interactable objects. We define a range of manipulation tasks using ManiSkill~\cite{taomaniskill3} as the simulator backend. Moreover, we use image augmentation to further reduce the visual gap between sim and real.

\paragraph{Reinforcement Learning.}
RL requires massive interaction between agents and the environments, usually with parallelism~\cite{taomaniskill3}. To better support RL, \acronym{} optimizes its implementation by only parallelizing 3DGS points that are linked with moving parts of the scene, \textit{i.e., robot $\mathcal{R}$ and objects $\mathcal{O}$}, and keeps other points cached with just one copy. This enables us to use a single GPU to run large parallelism to accelerate RL convergence.
\vspace{-5pt}
\section{Experiments}
\vspace{-5pt}
This section presents an empirical evaluation of \acronym{} to demonstrate its effectiveness. Our experiments are designed to address the following research questions:

\begin{itemize}[leftmargin=3mm]
    \item \textbf{Zero-shot Sim2real Imitation Learning.} Can \acronym{} effectively bridge the sim-to-real gap to enable zero-shot policy transfer? 
    \item \textbf{Closed-loop Policy Improvement.} Does access to a digital twin of the target deployment environment increase sampling efficiency and enable continual improvement after policy deployment via DAgger?
    \item \textbf{Visual Benchmarking.} Does performance in \acronym{} correlate with real-world performance? 
    \item \textbf{Virtual Teleoperation.} Does \acronym{} enable simulated data collection through human teleoperation?
    \item \textbf{Reinforcement Learning (RL).} Can \acronym{} narrow down the sim2real visual gap for visual RL?
\end{itemize}

\begin{figure}[h]
    \centering
    \includegraphics[width=0.98\linewidth]{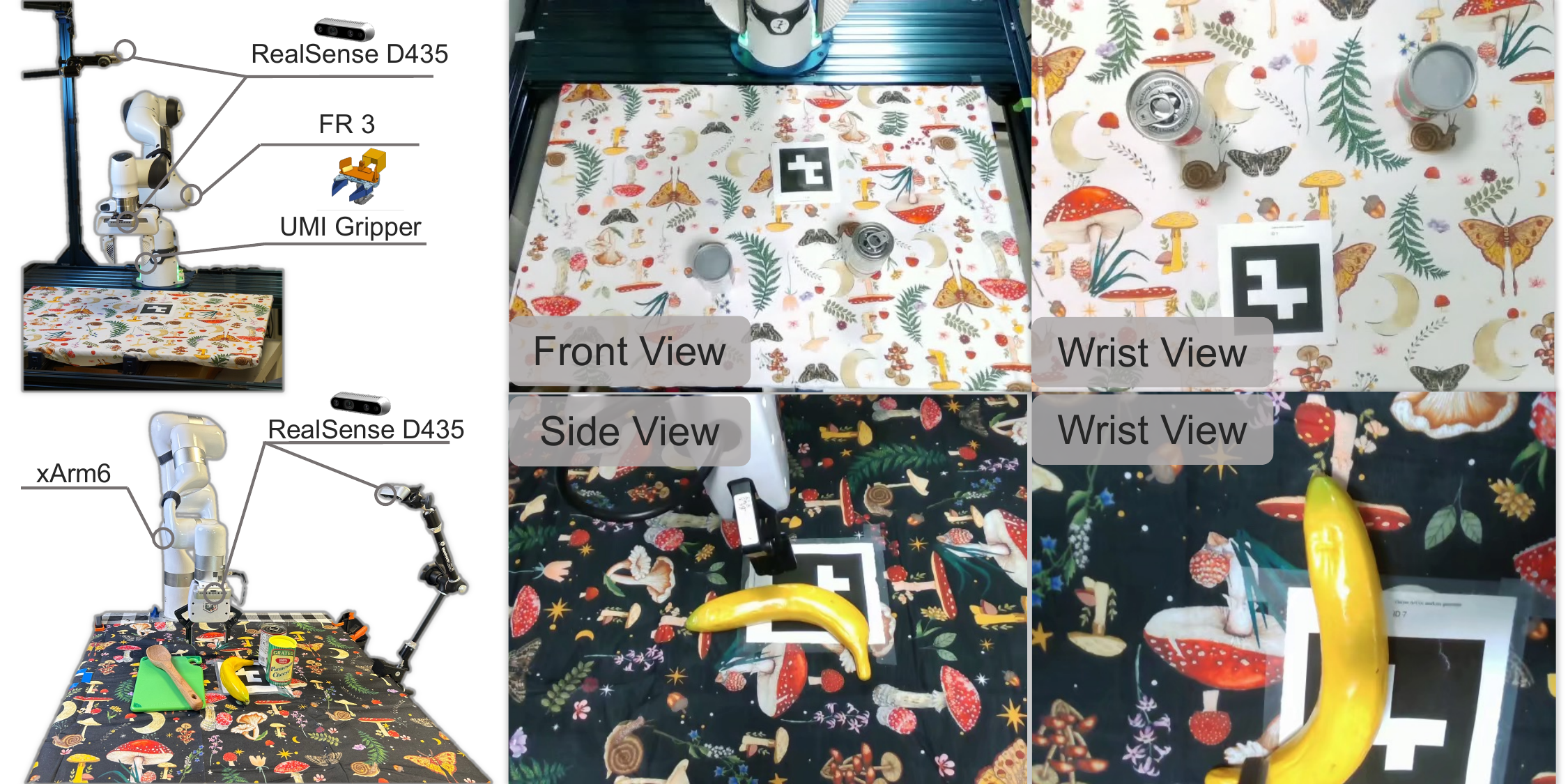}
    \caption{\small{\textbf{Real World Hardware Platforms.} For FR3, we set up two cameras: a third-person camera positioned in front of the robot (front view), and a wrist-mounted camera attached to the robot’s end effector (wrist view). For xArm6, we set up a third-person camera besides the robot (side view) and a wrist view.}}
    \label{fig:real-fr3}
    \vspace{-10pt}
\end{figure}

\paragraph{Hardware Platforms} Though \acronym{} scales to many robot embodiments, we consider three robot platforms for evaluation: a Franka Research 3 (FR3) robot with a UMI gripper~\cite{chi2024-umigripper}; a UF xArm6 with a parallel gripper; and a bimanual Galaxea R1 robot equipped with two 6-DoF arms. 

\paragraph{Experiment Protocol}
In total, we evaluate \acronym{} on 4 manipulation tasks on FR3 and 3 tasks on xArm. We also use R1 to demonstrate that \acronym{} supports virtual simulation teleoperation to collect data. For policy implementations, we use ACT~\cite{zhao2023-ACT} and Pi0~\cite{black2024-pi0} to show \acronym{} to be policy-agnostic. For visual benchmarking, we intentionally train policies with varying training settings and data sizes to evaluate the performance of `good' and `bad' policies. Please kindly refer to our released codebase for full implementation details due to limited space.

\begin{figure}[t]
    \centering
    \includegraphics[width=\linewidth]{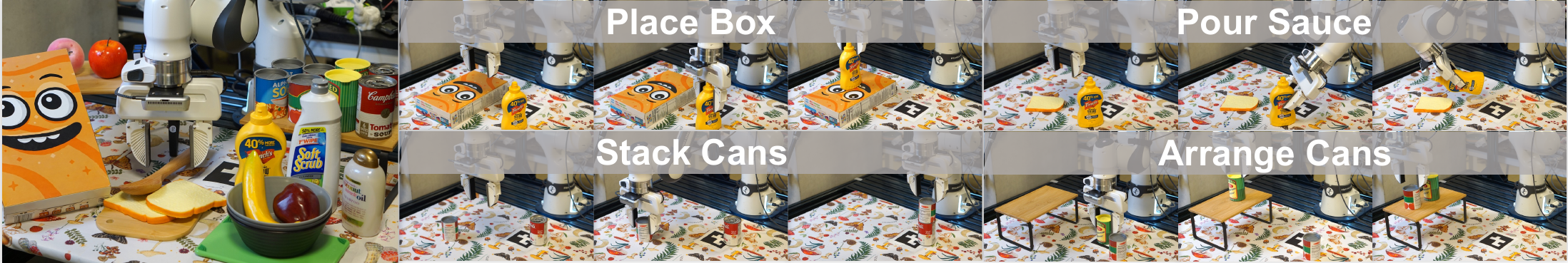}
    \caption{\textbf{FR3 Task Visualizations.} We design 4 real-world robotic tasks that involve distinct manipulation skills and diverse objects on FR3.}
    \label{fig:task-vis-fr3}
    \vspace{-10pt}
\end{figure}

\begin{figure}[t]
    \centering
\includegraphics[width=\linewidth]{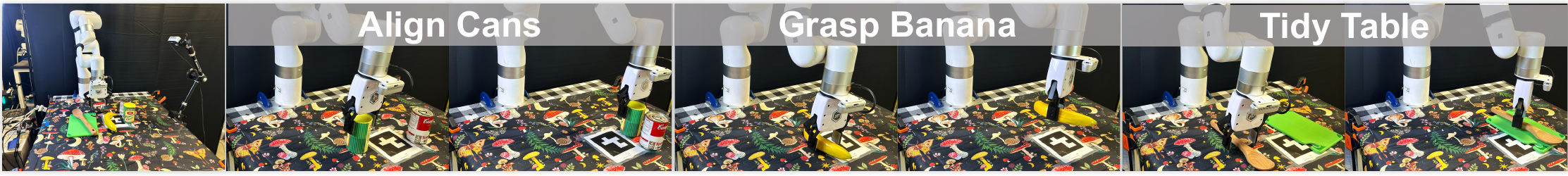}
\caption{\textbf{xArm Task Visualizations.} We design 3 manipulation tasks on xArm6.}
    \label{fig:task-vis-xarm}
    \vspace{-20pt}
\end{figure}

\paragraph{Manipulation Tasks Design}
We designed the following table-top manipulation tasks to evaluate performances, as shown in Fig.~\ref{fig:task-vis-fr3} and Fig.~\ref{fig:task-vis-xarm}:
\begin{itemize}[leftmargin=3mm]
    \vspace{-0.5mm}\item \underline{Place Box}. The bottle and the box are initialized randomly within a $45\,\text{cm} \times 45\,\text{cm}$ area. FR3 must pick up the bottle and place it onto the box.
    \vspace{-0.5mm}\item \underline{Pour Sauce}. The mustard bottle and the bread slice are put within a $45\,\text{cm} \times 45\,\text{cm}$ area. FR3 is required to pour the sauce onto the bread slice.
    \vspace{-0.5mm}\item \underline{Stack Cans}. Two cans are randomly placed within a $45\,\text{cm} \times 45\,\text{cm}$ area. FR3 must stack them.
    \vspace{-0.5mm}\item \underline{Arrange Cans}. Two cans are randomly placed within a $20\,\text{cm} \times 15\,\text{cm}$ area. A rack is randomly placed beside them. FR3 is tasked with placing one object on the rack, followed by placing the second object next to it.
    \vspace{-0.5mm}\item \underline{Align Cans}. xArm needs to grasp one can and put it next to the other can.
    \vspace{-0.5mm}\item \underline{Grasp Banana}. xArm should grasp the banana and rotate it by 30 to 60 degrees along the z-axis.
    \vspace{-0.5mm}\item \underline{Tidy Table}. xArm is required to pick up the kitchen spoon and place it onto the cutting board to clean the table.
\vspace{-1.5mm}
\end{itemize}

\begin{figure}[t]
    \centering
    \includegraphics[width=\linewidth]{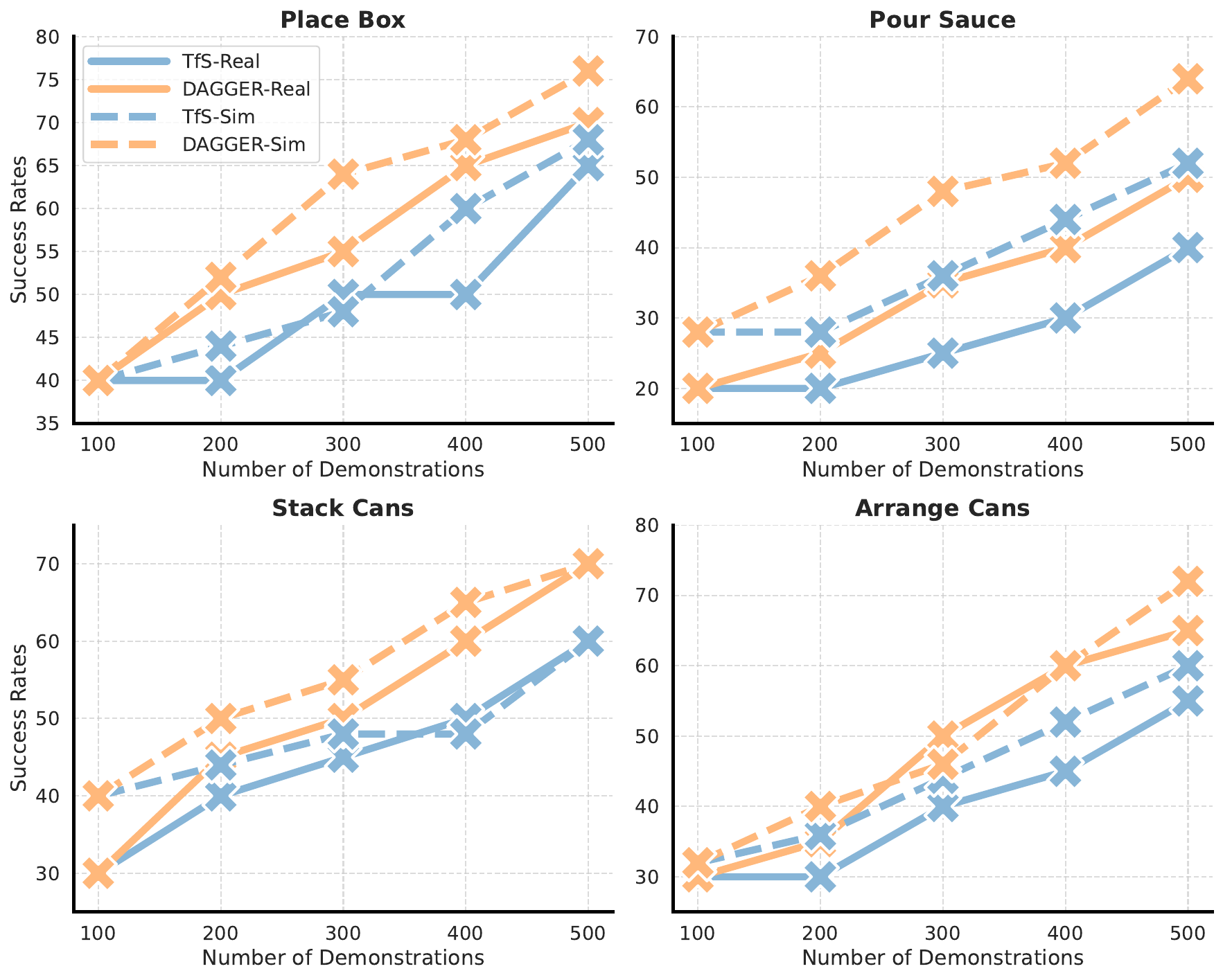}
    \caption{\textbf{Closed-loop Sim2real DAgger Training on FR3.} Policies are trained with sim data and deployed in both sim and real environments. DAgger consistently improves policy performances and outperforms training from scratch. }
    \label{fig:dagger}
    \vspace{-10pt}
\end{figure}

\begin{figure}[h]
    \centering
    \includegraphics[width=\linewidth]{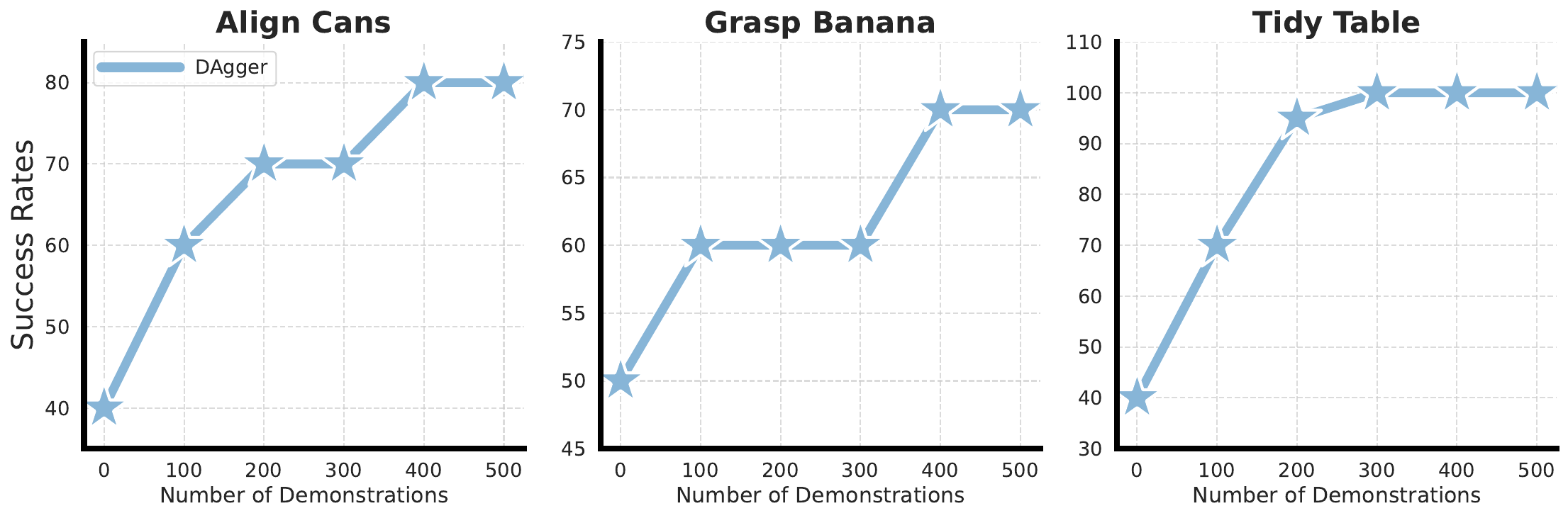}
    \caption{\textbf{Closed-loop Real2sim2real DAgger on xArm.} DAgger can also be used to improve real-world policies.}
    \label{fig:dagger-xarm}
    \vspace{-20pt}
\end{figure}

\subsection{Zero-shot Sim2real Imitation Learning}
Fig.~\ref{fig:dagger} shows the performances of policies trained with only simulation data. We leverage the MPlib motion planner~\cite{taomaniskill3}, utilizing predefined motions and poses. For each task, we collect 100 trajectories per iteration and train on the sum of all generated data. We can safely conclude that \acronym{} allows zero-shot sim2real policy transfer and shows promising success rates. More details are included in section~\ref{exp-sec:dagger}.

\subsection{Closed-loop DAgger for Continuous Policy Improvement}
\label{exp-sec:dagger}
The data collection uses same method as the zero-shot sim2real learning. For sim2real, we collect 100 expert trajectories for the initial iteration of DAgger training. In subsequent iterations, we evaluate the current policy and identify all failed trajectories. For each failure, we reset the environment to a preceding state where the task remains solvable, and collect additional corrective data using the motion planner starting from that state. This DAgger data collection process is repeated for four additional iterations, with each iteration generating 100 expert trajectories. As shown in Figure~\ref{fig:dagger}, our DAgger-based approach leads to significant performance improvements across all four tasks, compared to training from scratch (TfS) using only a supervised imitation learning objective.

DAgger can also be applied to improve real-world policies. Instead of starting from a randomly initialized policy, we can do real2sim2real DAgger learning. First we train a real-world ACT with a few demonstrations. Then we start from this checkpoint and do DAgger in \acronym{}. Fig~\ref{fig:dagger-xarm} shows that \acronym{} can improve performance after real-world policy deployment.

These results highlight the critical role of closed-loop DAgger Training. Leveraging our photo-realistic digital twin, researchers can collect essential corrective data that would be extremely difficult to obtain in the real world, primarily due to the challenges of precisely resetting objects to their pre-failure states.

\subsection{Benchmarking}

\begin{figure}[h]
    \centering
    \includegraphics[width=\linewidth]{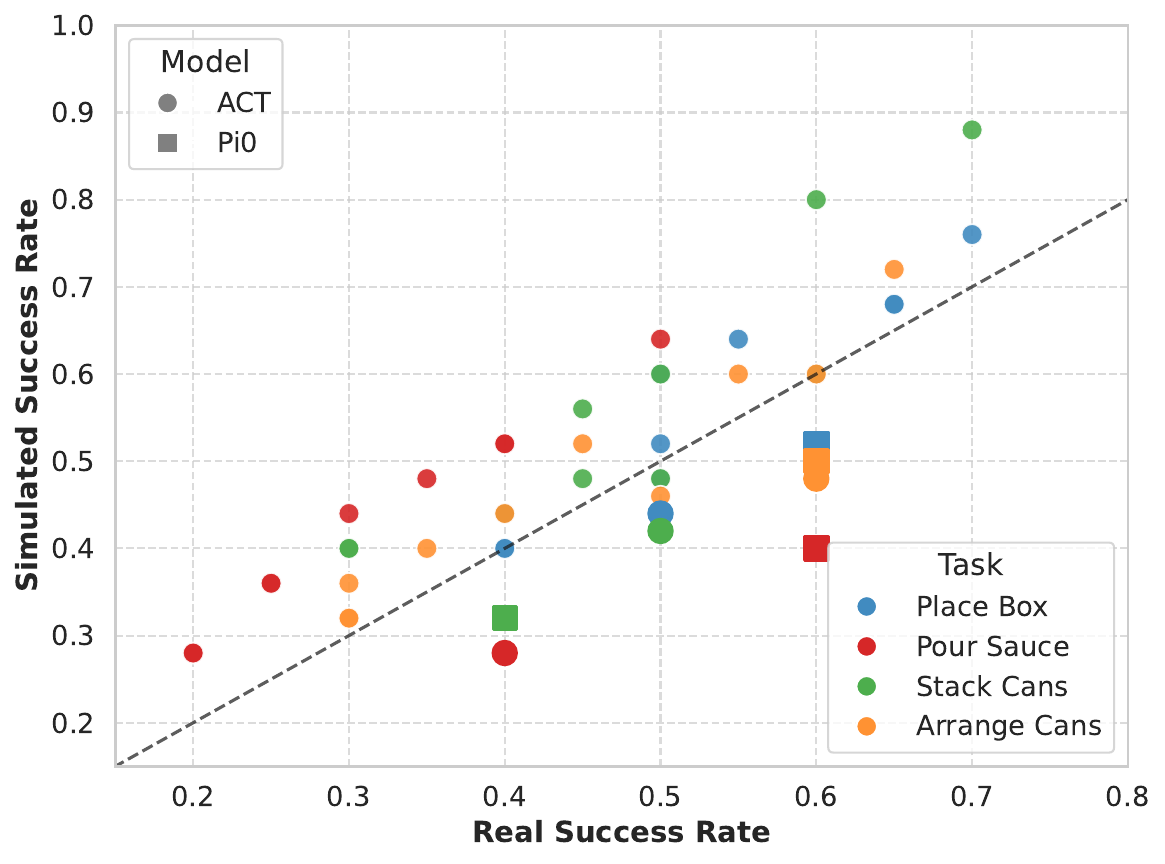}
    \caption{\textbf{Visual Benchmarking with FR3}. We roll out various policies in both real and sim. We intentionally use different policies and data sizes to show positive correlations of sim-real performance regardless of the quality of the policies.}
    \label{fig:benchmarking}
\end{figure}

In Fig.~\ref{fig:benchmarking}, we observe a strong correlation between simulation and real-world performance across all evaluated tasks and different policy architectures. This correlation indicates that \acronym{} can reliably predict real-world outcomes without requiring physical deployment of policies in the real world scene. As shown in Table~\ref{tab:benchmarking}, higher simulated performance consistently corresponds to higher success rates in real-world experiments. By leveraging the photorealistic rendering capabilities of 3DGS~\cite{kerbl20233d}, we establish a benchmarking framework that reflects real-world behaviors.

\begin{table}[t]
    \vspace{-5pt}
    \centering
    \scalebox{0.95}{
    {\fontsize{9pt}{11pt}\selectfont
    \begin{tabular}{lcccc}
        \toprule[1.2pt]
        & \multicolumn{2}{c}{\textbf{ACT}~\cite{zhao2023-ACT}} & \multicolumn{2}{c}{\textbf{Pi0}~\cite{black2024-pi0}} \\
        \cmidrule(lr){2-3} \cmidrule(lr){4-5}
        & real & sim & real & sim \\
        \midrule[1pt]
        \textbf{Place Box}    & 50.0\% & 44.0\% & 60.0\% & 52.0\% \\
        \textbf{Pour Sauce}   & 40.0\% & 28.0\% & 60.0\% & 40.0\% \\
        \textbf{Stack Cans}   & 50.0\% & 42.0\% & 40.0\% & 32.0\% \\
        \textbf{Arrange Cans} & 60.0\% & 48.0\% & 60.0\% & 50.0\% \\
        \textbf{Avg.}         & 50.0\% & 41.0\% & 55.0\% & 43.5\% \\
        \bottomrule
    \end{tabular}
    }}
    \vspace{5pt}
    \caption{\textbf{Visual Benchmarking of Real-world Policies on FR3.} We show the evaluation performance in both sim and real of policies trained with only real-world data.}
    \label{tab:benchmarking}
    \vspace{-20pt}
\end{table}

\subsection{Virtual Teleoperation}
\begin{figure}[h]
    \centering
    \vspace{-5pt}
    \includegraphics[width=\linewidth]{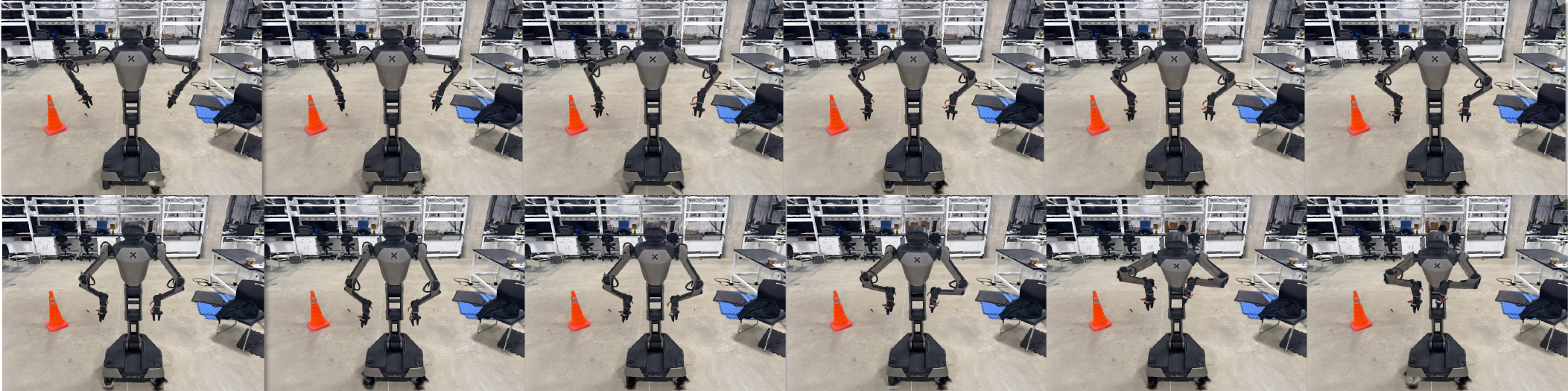}
    \caption{\textbf{Galaxea R1 Virtual Teleoperation}. We use a keyboard to teleop R1 and render photo-realistic videos.}
    \label{fig:r1-teleop}
    \vspace{-12pt}
\end{figure}

Simulation data can be used to scale up robot policy learning~\cite{huang2024diffusion, yang2025egovlalearningvisionlanguageactionmodels}. We show that through mouse and keyboard, we can collect teleoperation data in the simulation with real renderings, as illustrated in Fig.~\ref{fig:r1-teleop}.

\subsection{Visual Reinforcement Learning}
\begin{figure}[t]
    \centering
    \includegraphics[width=1.\linewidth]{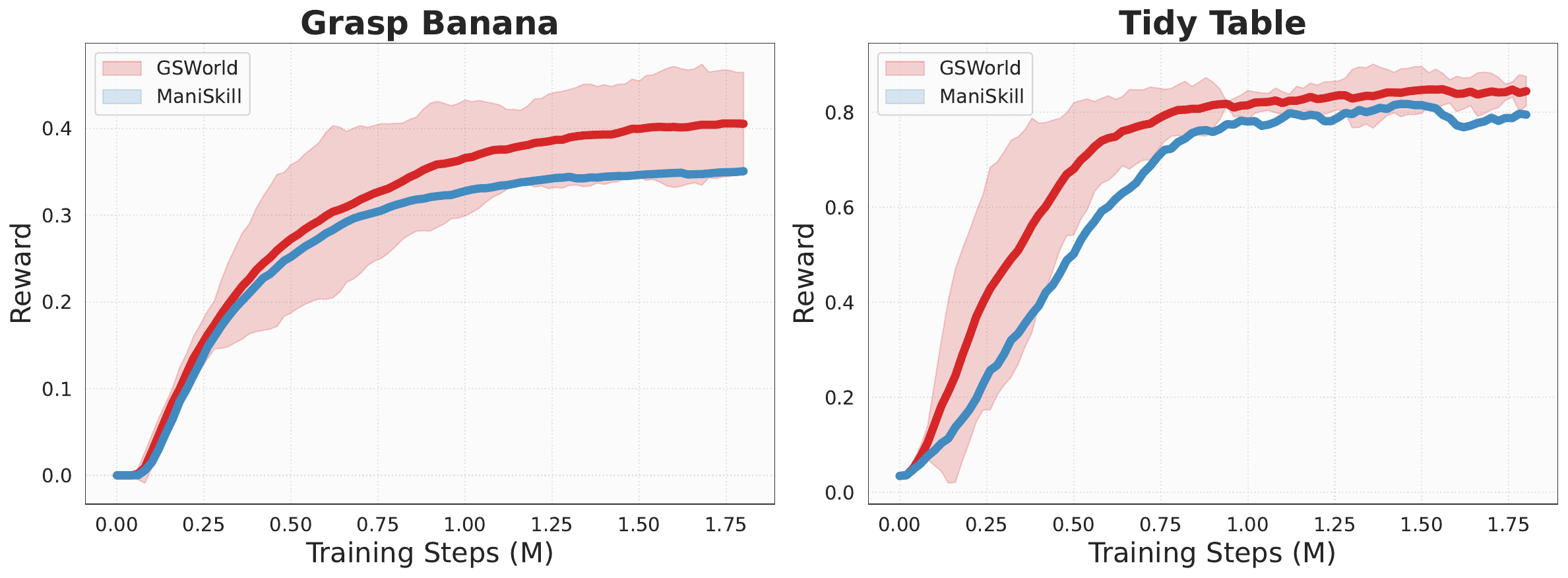}
    \caption{\textbf{SAC}. Results are aggregated over 3 runs. Directly training from ManiSkill is also plotted for comparison.}
    \label{fig:sac-rl}
    \vspace{-10pt}
\end{figure}

\acronym{} is designed to support parallel environments so that we can use this to learn visual RL policies, which has great potential in robot learning~\cite{huang2024diffusion, pan2025roboduet}. We trained asymmetric SAC~\cite{haarnoja2018soft} with \acronym{}, where the critic sees simulation-privileged information and the actor only uses robot joint position. As we aim to show that \acronym{} can reduce the visual gap for RL instead of acquiring good sim2real VRL policies, we train with no domain randomization except for color jittering for efficiency. We only use the third-person view since wrist camera shows significant gaps during RL exploration. Our real-world success rates for \textit{Grasp Banana} and \textit{Tidy Table} are 30\% and 20\%, while baseline ManiSkill reaches 0\% and 5\%, respectively. Training results are shown in Fig.~\ref{fig:sac-rl}.

\vspace{-5pt}
\section{Conclusion}
\vspace{-5pt}

In this work, we present a pipeline for constructing a photorealistic digital twin that delivers highly correlated performance metrics between simulation and real‐world deployments across different policy architectures and embodiments. Furthermore, we demonstrate that our environment can efficiently collect corrective data at scale, enabling more effective policy training.

\printbibliography

\clearpage


\appendix

\subsection{Neural Rendering Background}
\label{sec:appendix_rendering_math}

Point and surface splatting methods represent a scene explicitly via a mixture of 2D or 3D Gaussian ellipsoid. In the case of Gaussian Splatting, the geometry is represented as a collection of 3D Gaussian, each being the tuple \(\{\mathcal X, \mathbf\Sigma\}\) where $\mathcal{X} \in \mathbb{R}^3$ is the centroid of the Gaussian and $\Sigma$ is its covariance matrix in the world frame. This gives the probability density function%
\begin{equation}
    G(\mathcal{X}, \Sigma) = \exp{- \frac 1 2 \mathcal{X}^\top \Sigma^{-1} \mathcal{X}}\,.
\end{equation}%
Gaussian splatting decomposes it into a scaling matrix $\mathbf{S}$ and a rotation matrix $\mathbf{R}$ via \(\Sigma = \mathbf{R} \mathbf{S} \mathbf{S}^\top \mathbf{R}^\top\).  The color information in the texture is encoded with a spherical harmonics map \(\mathbf c_i = \mathrm{SH}_{\phi}(\mathbf d_i)\), which is conditioned on the viewing direction $\phi$. 

To optimize for features, existing methods tend to append an additional vector $\mathbf{f}_i \in \mathbb{R}^d$ to each Gaussian, which is rendered in a view-independent manner because the semantics of an object shall remain the same regardless of view directions. The rasterization procedure starts with culling the mixture by removing points that lay outside the camera frustum. The remaining Gaussians are projected to the image plane according to the projection matrix \(\mathbf{W}\) of the camera, which is then sorted from low to high using the distance from the virtual camera origin. This projection also induces the following transformation on the covariance matrix $\Sigma$:%
\begin{equation}
    \Sigma^{'} = \mathbf{J}\,\mathbf{W}\,\Sigma\,\mathbf{W^\top J^\top}\,,
\end{equation}
where $\mathbf{J}$ is the Jacobian of the projection matrix \(\mathbf{W}\). We can then render both the color and the visual features with the splatting algorithm:%
\begin{equation}
    \{\hat{\mathbf{F}}, \hat{\mathbf{C}\}} = \sum_{i \in N} \{\mathbf{f}_i, \mathbf{c}_i\}\cdot \alpha_i\, \prod_{j = 1}^{i - 1} (1 - \alpha_j)\,,
\end{equation}%
where $\alpha_i$ is the opacity of the Gaussian conditioned on $\Sigma^{'}$ and the indices $i\in N$ are in the ascending order determined by their distance to the camera origin.

Following the convention~\cite{qiu2024-featuresplatting}, \acronym{} assumes that per-gaussian feature vector $\mathbf{f}_i$ is isotropic. The rendered depth, images, and features are then supervised using L2 loss. 

\subsection{More Experimental Details}
\label{sec:appendix_exp_details}

\paragraph{Simulation Evaluation}
We evaluated each task across 25 random seeds in simulation. At the initiation of each episode, objects were randomly configured. A partial success criterion was established, awarding half credit for completing intermediate steps (e.g., successfully picking up an object or placing one can on the rack). Episodes for the \textit{placing box}, \textit{pouring sauce}, and \textit{stacking cans} tasks were limited to 500 time steps, while \textit{arranging cans} was allocated 800 time steps due to its extended task horizon. The control framework employed joint position control. The policy architecture accepted dual visual image inputs and the robot's proprioceptive data, subsequently generating target joint positions as action outputs.

\paragraph{Real-world Evaluation}
Real-world validation consisted of 10 experimental trials per policy, each with randomized initial conditions. Consistent with the simulation protocol, partial success metrics were implemented. The observational framework maintained parity with the simulation environment, utilizing visual inputs from two RealSense D435i cameras.

\paragraph{Policy Learning Implementation}
In our paper we use two policy architectures, i.e., ACT~\cite{zhao2023-ACT} and PI0~\cite{black2024-pi0}. For ACT, we adopted the original PyTorch implementation but replaced the visual backbone from ResNet-18 to DinoV2 ViT-S, similar to OpenTV~\cite{cheng2024-opentv}. We used 42 batch size, an SGD optimizer with a constant 0.0001 learning rate for 50\,000 iterations to train the policy. The rest of the training hyperparameters are consistent with the original ACT paper. For the PI0~\cite{black2024-pi0} model, we adopted the implementation on LeRobot. We used the publicly available PI0-base model weights for initialization. For the training, we freeze the visual and language backbone and train only the action expert. The optimization was done with an SGD optimizer with a constant 0.0001 learning rate for 50\,000 iterations with a batch size of 10.

\paragraph{DAGGER Detailed Performance}
In this section we provide the numerical results in Figure~\ref{fig:dagger}, which is shown in Table~\ref{tab:dagger}.

\begin{table*}
  \centering
  \setlength{\tabcolsep}{3pt}
  \scalebox{0.9}{%
    \fontsize{8pt}{11pt}\selectfont
    \begin{tabular}{ll *{10}{c}}
      \toprule[1.2pt]
      & & \multicolumn{5}{c}{\textbf{Place Box}} & \multicolumn{5}{c}{\textbf{Pour Sauce}} \\
      \cmidrule(lr){3-7} \cmidrule(lr){8-12}
      & & Iter 1 & Iter 2 & Iter 3 & Iter 4 & Iter 5
         & Iter 1 & Iter 2 & Iter 3 & Iter 4 & Iter 5 \\
      \midrule[1.2pt]
      \multirow{2}{*}{\textbf{sim}}
        & train from scratch 
          & 40\% & 44\% & 48\% & 60\% & 68\%
          & 28\% & 28\% & 36\% & 44\% & 52\% \\
        & dagger              
          & 40\% & 52\% & 64\% & 68\% & 76\%
          & 28\% & 36\% & 48\% & 52\% & 64\% \\
      \midrule
      \multirow{2}{*}{\textbf{real}}
        & train from scratch 
          & 40\% & 40\% & 50\% & 50\% & 65\%
          & 20\% & 20\% & 25\% & 30\% & 40\% \\
        & dagger              
          & 40\% & 50\% & 55\% & 65\% & 70\%
          & 20\% & 25\% & 35\% & 40\% & 50\% \\
      \addlinespace[0.8em]
      \toprule[1.2pt]
      & & \multicolumn{5}{c}{\textbf{Stack Cans}} & \multicolumn{5}{c}{\textbf{Arrange Cans}} \\
      \cmidrule(lr){3-7} \cmidrule(lr){8-12}
      & & Iter 1 & Iter 2 & Iter 3 & Iter 4 & Iter 5
         & Iter 1 & Iter 2 & Iter 3 & Iter 4 & Iter 5 \\
      \midrule[1.2pt]
      \multirow{2}{*}{\textbf{sim}}
        & train from scratch 
          & 40\% & 44\% & 48\% & 48\% & 60\%
          & 32\% & 36\% & 44\% & 52\% & 60\% \\
        & dagger              
          & 40\% & 56\% & 60\% & 80\% & 88\%
          & 32\% & 40\% & 46\% & 60\% & 72\% \\
      \midrule
      \multirow{2}{*}{\textbf{real}}
        & train from scratch 
          & 30\% & 40\% & 45\% & 50\% & 60\%
          & 30\% & 30\% & 40\% & 45\% & 55\% \\
        & dagger              
          & 30\% & 45\% & 50\% & 60\% & 70\%
          & 30\% & 35\% & 50\% & 60\% & 65\% \\
      \addlinespace[0.8em]
      \toprule[1.2pt]
      & & \multicolumn{10}{c}{\textbf{Avg.}} \\
      \cmidrule(lr){3-12}
      & & 
        \multicolumn{2}{c}{Iter 1} & \multicolumn{2}{c}{Iter 2} & 
        \multicolumn{2}{c}{Iter 3} & \multicolumn{2}{c}{Iter 4} & 
        \multicolumn{2}{c}{Iter 5} \\
      \midrule[1.2pt]
        \multirow{2}{*}{\textbf{sim}}
          & train from scratch 
            & \multicolumn{2}{c}{35\%} & \multicolumn{2}{c}{38\%} & 
              \multicolumn{2}{c}{44\%} & \multicolumn{2}{c}{51\%} & 
              \multicolumn{2}{c}{60\%} \\
          & dagger 
            & \multicolumn{2}{c}{35\%} & \multicolumn{2}{c}{46\%} & 
              \multicolumn{2}{c}{54.5\%} & \multicolumn{2}{c}{65\%} & 
              \multicolumn{2}{c}{75\%} \\
        \midrule
        \multirow{2}{*}{\textbf{real}}
          & train from scratch 
            & \multicolumn{2}{c}{30\%} & \multicolumn{2}{c}{32.5\%} & 
              \multicolumn{2}{c}{40\%} & \multicolumn{2}{c}{43.75\%} & 
              \multicolumn{2}{c}{55\%} \\
          & dagger 
            & \multicolumn{2}{c}{30\%} & \multicolumn{2}{c}{38.75\%} & 
              \multicolumn{2}{c}{47.5\%} & \multicolumn{2}{c}{56.25\%} & 
              \multicolumn{2}{c}{63.75\%} \\
      \bottomrule
    \end{tabular}
  }
  \vspace{10pt}
  \caption{Performance comparison between ACT and Pi0 methods with different training approaches (train from scratch vs.\ dagger) across various manipulation tasks and iterations (success rates in \%).}
  \label{tab:dagger}
\end{table*}

\subsection{Gaussian Splatting Training}
\label{sec:appendix_gs_training}
\paragraph{Implementation}
Our pipeline is compatible with any kind of gaussian splitting implementation. For simplicity, we use the official one~\cite{kerbl20233d} and their default code implementation without modifying anything. We take around 100 and 300 pictures for object and robot reconstruction, respectively. The model is trained with points initialized with Colmap~\cite{schonberger2016-colmap}, and saved at iteration 7000 and 30000.

\subsection{\acronym{} Data Collection}
\label{sec:appendix_data_gen}
\paragraph{Simulation}
We primarily rely on motion planning for data collection due to its fully automated nature. The task is decomposed into several subtasks, each associated with a specific motion. The robot executes these motions using \href{https://github.com/haosulab/MPlib}{MPlib}.

Our framework also supports data collection via teleoperation in simulation. We employ a built-in click-and-drag teleoperation system provided by ManiSkill~\cite{taomaniskill3}, which is operated using a keyboard and mouse. This system allows users to define keyframes and utilize a motion planner to achieve the desired robot pose.

\paragraph{Real World}
For real-world data collection, we utilize a VR-based teleoperation system. The interface includes an HTC Vive controller and base station, which track 6-DOF hand movements that are mapped to the robot arm's end-effector pose. Integration with the HTC Vive is achieved using the \href{https://github.com/TriadSemi/triad_openvr}{triad-openvr} package in conjunction with SteamVR. A position-velocity controller is implemented to ensure accurate tracking of the controller's transmitted poses. This setup enables efficient collection of demonstration trajectories, including both state and observation data. For the xArm6 platform, we use ACE-F~\cite{yan2025acef}.

\subsection{\acronym{} Usage example}
\label{sec:appendix_gsworld_wrapper}
Our pipeline is highly compatible with the gym interface. \acronym{} can be enabled by as few as one line of code, which add a wrapper layer on any existing environment. Below we show a code example to use \acronym{}.
\begin{lstlisting}[language=Python, frame=none, basicstyle=\small\ttfamily, commentstyle=\color{gray}\small\ttfamily,columns=fullflexible, breaklines=true, postbreak=\mbox{\textcolor{red}{$\hookrightarrow$}\space}, escapeinside={(*}{*)}]

    env = gym.make(
        env_id,
        robot_uids=args.robot_uid,
        obs_mode=args.obs_mode,
        control_mode=args.control_mode,
        render_mode=args.render_mode,
        reward_mode="dense",
        human_render_camera_configs=dict(shader_pack=args.shader),
        viewer_camera_configs=dict(shader_pack=args.shader),
        max_episode_steps=args.ep_len,
    )

    env = GSWorldWrapper(
        env=env,
        gs_cfg=args.gs_cfg,
        device="cuda",
    )

    # Then treat the env as the gym env.
    # Pixel observations are rendered with 3DGS.
    obs, _ = env.reset()
    obs, reward, terminated, truncated, info = env.step(action)
\end{lstlisting}

\end{document}